\pdfoutput=1
\documentclass[11pt]{article}

\usepackage[final]{acl}

\usepackage{subcaption}
\usepackage{times}
\usepackage{latexsym}
\usepackage{tabularx}
\usepackage[T1]{fontenc}
\usepackage[utf8]{inputenc}

\usepackage{microtype}

\usepackage{inconsolata}

\usepackage{graphicx}
\usepackage{booktabs}
\usepackage{multirow}
\usepackage{bm}
\usepackage{amsmath}
\title{GAProtoNet: A Multi-head Graph Attention-based Prototypical Network for Interpretable Text Classification}

\author{
 \textbf{Ximing Wen\textsuperscript{1}},
 \textbf{Wenjuan Tan\textsuperscript{2}},
 \textbf{Rosina O. Weber\textsuperscript{1}}
\\
\\
 \textsuperscript{1}College of Computing and Informatics, Drexel University, Philadelphia, USA\\
 \textsuperscript{2}Department of Computer Science, Tsinghua University, Beijing, China
\\
 \texttt{\{xw384, rw37\}@drexel.edu} $\qquad$ \texttt{tanwj21@mails.tsinghua.edu.cn}
}

\begin{document}
\maketitle
\begin{abstract}
Pretrained transformer-based Language Models (LMs) are well-known for their ability to achieve significant improvement on text classification tasks with their powerful word embeddings, but their \textit{black-box} nature, which leads to a lack of interpretability, has been a major concern. In this work, we introduce GAProtoNet, a novel \textit{white-box} Multi-head Graph Attention-based Prototypical Network designed to explain the decisions of text classification models built with LM encoders. In our approach, the input vector and prototypes are regarded as nodes within a graph, and we utilize multi-head graph attention to selectively construct edges between the input node and prototype nodes to learn an interpretable prototypical representation. During inference, the model makes decisions based on a linear combination of activated prototypes weighted by the attention score assigned for each prototype, allowing its choices to be transparently explained by the attention weights and the prototypes. Experiments on multiple public datasets show our approach achieves superior results without sacrificing the accuracy of the original black-box LMs. We also compare with four alternative prototypical network variations and our approach achieves the best accuracy and F1 among all. Our case study and visualization of prototype clusters also demonstrate the efficiency in explaining the decisions of black-box models built with LMs. Our codes are available at \url{https://github.com/ximingwen/GAProtoNet}.
\end{abstract}

\section{Introduction}
Deep learning models, especially transformer-based Language Models (LMs) such as BERT \cite{devlin2018bert}, RoBERTa \citep{liu2019roberta} have significantly contributed to advancements in NLP, especially text classification. However, despite their \textit{state-of-art} performance, their complexity and \textit{black-box} nature obscure the decision-making process and hinder their interpretability. Now more and more real-world applications also desire classification models built with LMs to be interpretable, since it allows end-users to comprehend the decision-making process, fostering trust and encouraging adoption. To address this need, there is growing interest in enhancing the interpretability of text classification models built with LMs.

Recent efforts have focused on redesigning neural networks to be inherently interpretable, based on the classic framework of prototypical learning (\citealp{datta1995learning}). These models actively learn prototype vectors hrough training, which are representative cases from previous observations, to explain decisions more intuitively. This methodology is first introduced into image domain by \citet{li2018deep} and \citet{chen2019looks},  and then applied in text classification field with different structure variations \citep{ming2019interpretable, hong2020protorynet, plucinski2021prototypical, das2022prototex}. However, despite this \textit{white-box} framework being efficient in training and improving the model's intrinsic interpretability, there is still some performance gap compared with the original black-box models.

On the other hand, there exist strategies such as graph attention network (GAT) \cite{velickovic2017graph}, that is known for its ability to capture the importance of neighboring nodes in a graph through attention mechanisms, enabling more effective and expressive feature representation learning for each node. This approach allows GAT to dynamically assign different weights to different neighbors, which enhances the model's performance on various graph-based tasks\citep{ wang2019kgat, xie2020mgat, bhatti2023mffcg}. This inspires us to analogize learning the relatedness between the input vector and the prototype vectors as constructing edges and learning their weights between nodes. 

To address the challenge of the performance gap, we propose a novel \textit{white-box} Multi-head \textbf{G}raph \textbf{A}ttention-based \textbf{Proto}typical \textbf{Net}work (\textbf{GAProtoNet}) designed to explain the decisions of text classification models built with LM encoders. Our approach incorporates a prototype layer on top of a fine-tuned LM and utilizes multi-head graph attention \cite{velickovic2017graph} to efficiently learn an interpretable prototypical representation by selectively constructing edges between encoded representations and their neighboring prototypes. In the reference time, the decision is solely based on a linear combination of prototypes weighted by the attention scores assigned by attention heads. Our model archives superior performance without sacrificing the accuracy of the original LM while facilitating transparent decision-making progress. The novelty of our work lies in being the first to represent prototypes and input vectors as nodes within graphs, and actively train prototype vectors utilizing edges constructed by graph attention for interpretable text classification. The contribution of our work can be summarized as follows:
\begin{itemize}
    \item 
    We propose a new prototypical framework that leverages multi-head graph attention to selectively construct edges between input and prototypes which indicates relatedness. Our model is inherently interpretable and the performance is superior without sacrificing the accuracy of the original models.
    \item We did extensive comparison experiments with variations of prototype-based networks on five public benchmark datasets, including binary, four-label, ten-label classification. Our approach achieves the best accuracy and F1. 
    \item We evaluate the interpretability of GAProtoNet through various design criteria and demonstrate that the explanations provided by GAProtoNet are of high quality.

\end{itemize}
The next section describes our proposed approach. \S \ref{3} presents experiments, while \S \ref{4} presents an analysis of the interpretability exhibited in the proposed approach. We describe related work in \S \ref{5}. We conclude in \S \ref{6} where we reiterate our contribution. Limitations and future work are pointed out in \S \ref{7}.

\section{Graph-Attention ProtoNet}\label{2}

\begin{figure*}[t]
\centering
\includegraphics[width=\textwidth]{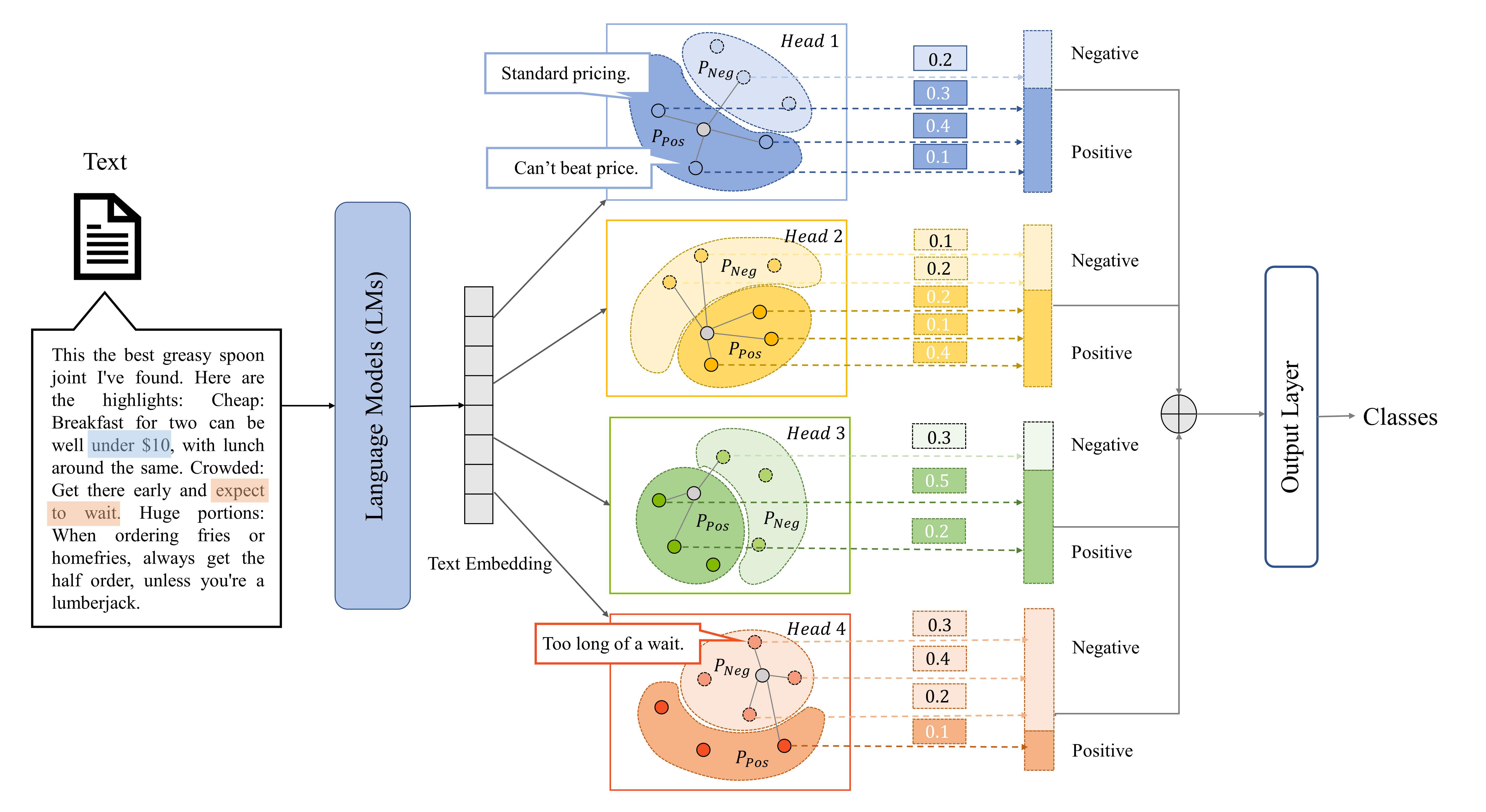}
\caption{\textbf{Overview of the GAProtoNet Architecture}: The GAProtoNet architecture consists of three primary components: a Text Embedding Layer using LMs, a Prototype layer, and Graph Attention. The text embeddings are linearly transformed to produce query vectors. Each attention head will construct a graph based on the attention score between the query vectors and predefined prototypes. As shown in the figure, different heads will activate prototypes in different semantic aspects, assigning different attention score to negative ones and positive ones. An interpretable prototypical representation will be formed with a linear combination of all prototypes weighted by the attention score and sent to the output layer for classification.}
\label{fig:example}
\end{figure*}

Figure \ref{fig:example} illustrates the overall architecture of GAProtoNet, which consists of three major components: (1) \textit{Text Embedding based on Language Models (LMs)}, responsible for converting input text into high-dimensional vectors that capture its semantic information; (2) \textit{Prototype Layer}, responsible for forming multiple typical prototype vectors that encapsulate distinct semantic aspects of the input text as well as enhancing the interpretability of the model; and (3) \textit{Graph Attention Mechanism}, which efficiently learn the relatedness between the text embedding vector and its neighboring prototype vectors, which are later used to form a prototypical representation of the text and passed to output layer. Detailed descriptions for each component are presented in the following subsections.

\subsection{Text Embedding encoded by Language Models (LMs)} Given the remarkable performance of pre-trained language models  RoBERTa, XLNet, and DistilBERT on a wide range of NLP tasks, we leverage their capabilities for efficient text embeddings. These models have been trained on vast text datasets and can be fine-tuned for specific downstream tasks. By utilizing these pre-trained language models, we embed input text into high-dimensional vectors, which serve as representations that capture the deep semantic information of the text:
\begin{equation}  
\bm{s}=\textbf{LM}(\bm{x})  
\end{equation}
where $\bm{x}$ denotes the input text. $\bm{s}$ is the semantic information representation vector.

\subsection{Prototype Layer} \label{prototype layer}
In the prototype layer, we define $M$ prototype vectors $\bm{P} = \{\bm{p}_j\}_{j=1}^M$  that represent typical features in the vector space of training data. For each semantic representation vector passed from the text embedding layer, we compute the attention score vector indicating the relatedness between itself and all prototype vectors with a graph attention model introduced in \S \ref{graph attention}. 

Note that these prototype vectors are randomly initialized. They are learned through active training with loss defined in \S \ref{2.4}. 
Their representativeness is improved by updating the weights during each training epoch, resulting in an inherently interpretable classifier.

\subsection{Graph Attention Mechanism} \label{graph attention}
The graph attention mechanism in our model captures the relatedness between the embedding text vector and the prototype vectors, allowing for the derivation of prototypical representations for the embedding vectors mentioned in \S \ref{prototype layer}. To effectively capture diverse semantic aspects and intricate patterns in the data, we utilized a multi-head approach to comprehensively model this relatedness. The query nodes in the graph represent the input text embedding vector while the key nodes represent the prototype vectors. Edges and weights are constructed based on the following approach:
 
\paragraph{Single-Head Linear Transformation} For each text embedding input $\bm{s}$, we utilize matrix $\bm{W}^q$ to transform it into query vector $\bm{q}$. For each prototype $\bm{p}_j$, we utilized matrix $\bm{W}^k$ to transform it into $\bm{k}_j$:
\begin{equation}
\bm{q} = \bm{W}^q \bm{s},   \bm{k}_j = \bm{W}^k \bm{p}_j
\end{equation}

\paragraph{Multi-Head Linear Transformation} Now instead of using one single set of weight,  we utilize a set of learnable matrix linear transformations $\left\{\bm{W}^q_i\right\}_{i=1}^H$ and $\left\{\bm{W}^k_i\right\}_{i=1}^H$ to generate a series of query nodes $\left\{\bm{q}_i\right\}_{i=1}^H$ and prototypes $\left\{\bm{k}_{ij}\right\}_{i=1}^H$ with $H$ representing the number of attention heads. For each head $i$, we have:

\begin{equation} 
\bm{q}_i = \bm{W}^q_i \bm{s}, \bm{k}_{ij} = \bm{W}^k_i \bm{p}_j
\end{equation}

\paragraph{Attention Score Computation} After the transformation with each head $i$, we calculate the similarity between the node $\bm{q}_i$ and the nodes  $\left\{\bm{k}_{ij}\right\}_{j=1}^M$
based on the dot product:

\begin{equation}
sim(\bm{q}_i,\bm{k}_{ij}) = dot(\bm{q}_i \bm{k}_{ij})/d_k
\end{equation}

where $d_k$ denotes the dimension of key node vector $k$.

The attention scores are then derived by applying the sigmoid activation function $\sigma$  to the similarity vector:

\begin{equation}
\alpha_{ij} = \sigma(sim(\bm{q}_i,\bm{k}_{ij}))
\end{equation}
This attention score represents the weight of each prototype $p$ in shaping the decision-making process. As shown in Figure 1, our experimental results demonstrate that each attention head tends to activate prototypes corresponding to distinct semantic aspects. Moreover, the attention scores assigned to negative and positive prototypes suggest a bias towards a negative or positive final classification, respectively.
\paragraph{Graph Edge Construction}  For each attention head $i$, a graph is constructed by activating edges between the transformed prototype nodes $\bm{k}$ and text embedding node $\bm{q}$ when attention scores exceed a certain threshold \( \tau \). For each attention head $i$, the edge between the transformed prototype $\bm{q}$ and transformed $j$-th prototype node $\bm{k}_j$ can be expressed as:
\begin{equation}
E = \left\{ (\bm{q}_i, \bm{k}_{ij}) \mid \alpha
_{ij} > \tau \right\}
\label{e}
\end{equation}

\paragraph{Interpretable Prototypical Representation} The prototypical representation vector for the text embedding vector $s$ is formed by computing a linear combination of the neighboring prototype nodes from all attention heads weighted by the attention score. 

For each attention head $i$, we first normalize the attention scores for neighboring transformed prototype nodes of $\bm{q}_i$. Thus, for each prototype node $\bm{k}_j$ under attention head $i$, its normalized score is:

\begin{equation}
\gamma_{ij} = \frac{\alpha_{ij}}{\sum_{k \in \mathcal{N}(\bm{q}_i)} \alpha_{ik}}
\end{equation}

The prototypical representation vector $\bm{r}$ for $\bm{s}$ under attention head $i$ is computed as:

\begin{equation}
\bm{r}_i = \sum_{j \in \mathcal{N}(\bm{q}_i)} \gamma_{ij} \bm k_{ij}
\end{equation}
where  $N (\bm{q}_i)$ is the set of neighboring nodes for node $\bm{q}$ under attention head $i$ constructed under the definition of $E$ in \ref{e}.

\paragraph{Output Layer} The interpretable prototypical representations obtained from all heads are concatenated to form a single vector. This vector is then utilized as input for the output layer to perform text classification. 

\subsection{Prototype Projection}
To understand the natural language meaning of each prototype vector \( \bm{p}_j \), we match each prototype with the sample text embedding vector \( \bm{s}_j \) in the dataset \( \mathcal{D} \) that has the highest similarity, and assign that sample \( \bm{x}_j \) as the prototype text. Let \( \mathcal{D} \) represent the training dataset, then:

\begin{equation}
\textbf{Text of } \bm{p}_j \leftarrow \arg \max_{\substack{\bm{x}_j \in \mathcal{D}}} sim(\bm{s}_j, \bm{p}_j)
\end{equation}

where \( sim(\bm{s}_j, \bm{p}_j) \) denotes the similarity measure between the sample text \( \bm{s}_j \) and the prototype vector \( \bm{p}_j \).

During text classification tasks, we observe the graph model formed between embedded text and prototypes, as well as the edge weights, to interpret the classification process.

\subsection{Training Objective } \label{2.4}

In this study, our designed composite loss function consists of three key components: \textit{Accuracy Loss}, \textit{Proximity Loss}, and \textit{Diversity Loss}.

\paragraph{Accuracy Loss}

We employ the cross entropy loss to guide the model training. This loss measures the discrepancy between the predicted probability distribution and the true labels, optimizing the model to assign higher probabilities to the correct classes. Specifically, the accuracy loss is defined as:
\begin{equation}
\mathcal{L}_{Acc} = - \sum_{i=1}^{N} \sum_{j=1}^{C}y_{ij} \log(\hat{y}_{ij}),
\end{equation}
where $N$ is the number of samples, while $C$ is the number of classes. $y_{ij}$ is the true one-hot label of sample $i$ for class $j$, while $\hat{y}_{ij}$ is the predicted probabilities.

\paragraph{Proximity Loss}

To ensure that each prototype can be projected onto a similar sample in the training data, we introduce the Proximity Loss. This loss measures the distance between prototypes and data points, penalizing prototype-sample pairs that are far apart. We use Euclidean distance as the distance metric and calculate the minimum distance between each prototype and all samples. The average of the minimum distances across all prototypes is then taken as the Proximity Loss:
\begin{equation}
\mathcal{L}_{Prox} = \frac{1}{M} \sum_{j=1}^{M} \min_{i} || \bm{p}_j - \bm{s}_i ||_2^2
\end{equation}

where $M$ is the number of prototypes,  $\bm{p}_j$is the vector representation of the $j$-th prototype, and $\bm{s}_i$ is the embedded vector representation of the $i$-th sample in the training set.

\paragraph{Diversity Loss}

To encourage diversity among prototypes and avoid redundancy, we introduce the Diversity Loss. This loss aims to encourage prototypes to be distributed as diversely as possible in the feature space. We achieve this by penalizing the average distance between all pairs of prototypes:
\begin{equation}
\mathcal{L}_{Div} = -\frac{1}{M(M-1)} \sum_{j=1}^{M} \sum_{k \neq j} || \bm{p}_j - \bm{p}_k ||_2
\end{equation}
% The negative sign is used to convert the objective of maximizing the distance into a minimization problem for the loss function.

\paragraph{Composite Loss Function}

By combining the above three loss functions, we obtain the composite loss function:
\begin{equation}
\mathcal{L} = \lambda_1 \cdot \mathcal{L}_{Acc} + \lambda_2 \cdot \mathcal{L}_{Prox} + \lambda_3 \cdot \mathcal{L}_{Div}
\end{equation}
where $\lambda_1$, $\lambda_2$, and $\lambda_3$ are hyperparameters that balance the weights between different loss terms. These hyperparameters can be determined experimentally to optimize the model's performance.

\begin{table*}[t]
  \centering
    \fontsize{8}{10}\selectfont  % Set the font size to 8pt with a line spacing of 10pt
\begin{tabularx}{\textwidth}{l|*{3}{>{\centering\arraybackslash}X}|*{3}{>{\centering\arraybackslash}X}|*{3}{>{\centering\arraybackslash}X}|*{3}{>{\centering\arraybackslash}X}|*{3}{>{\centering\arraybackslash}X}}
    \specialrule{1.5pt}{0pt}{1pt}
    \multicolumn{1}{l}{} & \multicolumn{3}{|c|}{\textbf{Hotel}} & \multicolumn{3}{c|}{\textbf{IMDb}} & \multicolumn{3}{c|}{\textbf{Yelp}} & \multicolumn{3}{c|}{\textbf{Tweet}} & \multicolumn{3}{c}{\textbf{Yahoo}} \\
    \midrule
    & \textbf{Acc.} & \textbf{Rec.} & \textbf{F1} & \textbf{Acc.} & \textbf{Rec.} & \textbf{F1} & \textbf{Acc.} & \textbf{Rec.} & \textbf{F1} & \textbf{Acc.} & \textbf{Rec.} & \textbf{F1} & \textbf{Acc.} & \textbf{Rec.} & \textbf{F1} \\
    \midrule
    DistilBERT & 97.49 & 98.22 & 97.57 & 92.91 & 93.53 & 92.93 & 97.25 & 97.51 & 97.43 & 76.84 & 76.84 & 76.89 & 71.17 & 71.17 & 71.97 \\
    XLNet & 98.73 & 98.21 & 98.77 & 95.79 & 96.71 & 95.82 & 98.12 & 98.25 & \textbf{98.40} & 79.52 & 79.52 & 79.26 & \textbf{75.80} & \textbf{75.80} & \textbf{75.45} \\
    RoBERTa-large & 98.97 & 99.33 & 99.00 & 96.38 & \textbf{97.08} & 96.38 & 98.18 & 98.57 & 98.21 & 81.63 & 81.70 & 81.63 & 75.46 & 75.46 & 74.94 \\
    \midrule
    ProSeNet & 92.17 & 93.72 & 93.06 & 86.69 & 87.84 & 87.53 & 92.58 & 91.24 & 92.17 & 73.24 & 72.35 & 73.54 & 68.17 & 68.23 & 68.19 \\
    ProtoryNet & 94.45 & 95.59 & 94.55 & 88.21 & 87.14 & 88.45 & \underline{95.32} & 94.08 & \underline{94.62} & 72.90 & 72.85 & 72.28 & \underline{70.62} & 69.03 & \underline{70.58} \\
    ProtoCNN & 94.32 & 95.49 & 94.34 & 89.32 & 91.03 & 89.65 & 87.25 & 88.47 & 89.98 & 64.32 & 63.51 & 63.36 & 66.72 & 65.98 & 64.21 \\
    ProtoTEX & \underline{95.90} & \underline{96.37} & \underline{96.23} & \underline{92.39} & \underline{93.21} & \underline{91.36} & 93.07 & \underline{94.40} & 94.26 & \underline{75.28} & \underline{75.14} &\underline{ 75.20} & 70.32 & \underline{70.33} & 70.33 \\
    \midrule
    \multicolumn{16}{c}{\textsc{Single-head GAProtoNet Variations}} \\
    \midrule
    DistilBERT + SG & 97.14 & 96.67 & 97.20 & 92.79 & 93.73 & 92.82 & 97.38 & 97.76 & 97.40 & 75.72 & 75.72 & 76.00 & 70.67 & 70.67 & 71.35 \\
    XLNet + SG & 98.62 & 98.88 & 99.11 & 95.36 & 94.65 & 95.73 & 97.65 & \textbf{98.70} & 98.21 & 79.80 & 79.80 & 79.76 & 73.77 & 73.77 & 73.65 \\
    RoBERTa + SG & 98.94 & 98.67 & 98.89 & 96.35 & 96.97 & 96.23 & 98.14 & 98.11 & 98.17 & 81.42 & 81.40 & 81.42 & 72.23 & 72.23 & 71.16 \\
    \midrule
    \multicolumn{16}{c}{\textsc{Multi-head GAProtoNet Variations}} \\
    \midrule
    DistilBERT + MG & 97.94 & 97.78 & 98.00 & 92.79 & 94.06 & 92.86 & 97.40 & 96.93 & 97.43 & 75.43 & 75.40 & 75.40 & 71.40 & 71.40 & 72.00 \\
    XLNet + MG & 99.08 & 99.33 & 99.11 & 95.96 & 95.96 & 95.96 & 98.19 & 98.69 & 98.23 & 80.64 & 80.64 & 80.48 & 74.12 & 74.12 & 74.74 \\
    RoBERTa + MG & \textbf{99.09} & \textbf{99.56} & \textbf{99.12} & \textbf{96.53} & 96.85 & \textbf{96.53} & \textbf{98.27} & 98.07 & 98.39 & \textbf{81.84} & \textbf{81.81} & \textbf{81.85} & 73.79 & 73.79 & 73.95 \\
    \midrule
    Improvement (\%) & 3.19 & 3.19 & 2.89 & 4.14 & 4.46 & 5.17 & 2.95 & 4.29 & 3.77 & 6.56 & 6.67 & 6.65 & 3.50 & 3.79 & 4.16 \\
      \specialrule{1.5pt}{0pt}{1pt}
\end{tabularx}

\caption{Accuracy, recall and F1 scores for baseline models and GAProtoNet variations across three datasets. 'SG' denotes single-head GAProtoNet, while 'MG' represents multi-head GAProtoNet. Bolded values indicate the highest scores in each column across all sections, while underlined values highlight the best performance within the prototypical baseline variation section. The percentage improvement reflects the difference between the highest performance of GAProtoNet and that of other prototype-based variations.} 
\label{tab:results1}
\end{table*}

\section{Performance Experiments} \label{3}
In this section are discussed the datasets used in the experiments, and the models used as black-box baselines, prototypical baselines, and the variations of the proposed approach. The overall hypothesis is that the proposed approach performs on par with black-box and outperforms prototype baselines in the dataset used for text classification. 
\subsection{Datasets and Metric}

We evaluate our approach on three binary public benchmark datasets: \textbf{Hotel Reviews} \footnote{ https://www.kaggle.com/datafiniti/ hotel-reviews.}prepared by \citep{hong2020protorynet}, \textbf{IMDb} \footnote{https://huggingface.co/datasets/stanfordnlp/imdb.} \citep{maas-EtAl:2011:ACL-HLT2011} and \textbf{Yelp Polarity Reviews}. All of them are balanced datasets. To demonstrate our model's performance on more challenging tasks, we also evaluate it on \textbf{Tweet} \cite{mohammad-etal-2018-semeval} and \textbf{Yahoo} \cite{zhang2015character}, which are 4-class and 10-class datasets seperately.

\paragraph{Metric} We use accuracy, recall and F1-Score as metrics to evaluate models’ performance. For each model, we run 5 times and calculate the average as the final reported results.

\subsection{Models and Settings} 
We select three \textit{black-box} LM as baselines and four different prototype-based variations for comparison experiments. We also test two variations of GAProtoNet, one with a single attention head and another with attention heads $=4$. We train each model with one single NVIDIA GTX 3090 or GTX 4090. We use Adam as the optimizer and the learning rate is  $1e-4$. Due to the limitation of GPU RAM, we choose a batch size of 4 and an accumulated gradient step of 64.
\paragraph{LM Baselines} We select the following pre-trained \textit{black-box} language model without prototypes as the powerful baselines: \textbf{DistilBERT \cite{sanh2019distilbert}}, \textbf{RoBERTa-large \cite{liu2019roberta}} and \textbf{XLNet \cite{yang2019xlnet}}. For all the three LMs, we use the same simple MLP (multilayer perceptron) with two hidden layers over the output of the \texttt{CLS} token from the last hidden layer of the LM encoder for classification.

\paragraph{Prototype-based Networks Variations} We select the following 4 prototyical network variations for comprehensive comparison experiments against our GAProtoNet:\textbf{ProSeNet \cite{ming2019interpretable}}, \textbf{ProtoryNet \cite{hong2020protorynet}}, \textbf{ProtoCNN \cite{plucinski2021prototypical}} and  \textbf{ProtoTEX \cite{das2022prototex}}.

% We test ProtoCNN and ProtoTEX with RoBERTa as the encoder for comparison.

\subsection{Experiment Result}

%   four prototypical DL models without GA with max performance 95.9, 96.37, and 96.23 for accuracy, precision, and F1. It also shows three DL models that are not interpretable with max performance for acc being 98.97. Then, it shows two variations of the GA interpretable prototypical deep learning architecture (single- and multi-head) where single-head reaches 98.94 and multi-head reaches 99.09 for the Hotel dataset. This performance is consistent throughout three different datasets.

\paragraph{Binary Classification } Thr first three section in Table \ref{tab:results1} summarizes our evaluation results on binary classification. We observe that GAProtoNet either closely matches or exceeds the performance of its baseline LLM across our experiments, supporting GAProtoNet's interpretability is not achieved at the cost of performance. Within four types of prototypical variations, ProtoTEX and ProtoryNet have the best performance for accuracy, recall, and F1 for all three datasets, but GAProtoNet still improves 3\%-6\% upon their performance. This indicates that leveraging graph attention to construct edges between input embeddings and prototypes could be more efficient than heuristic distance measurement.  For the variations of GAProtoNet, multi-head GAProtoNet have better performance compared with single-head, with the RoBERTa as the encoder showing the best results. This could be due to that the multi-head attention enables the network to capture semantic meanings from different perspectives, thus potentially improving the network's performance.

\paragraph{Extend to Multi-label Classification} The last two section in Table \ref{tab:results1} summarizes our evaluation results on multi-label classification. For the 4-class dataset, our model is the best overall. For the 10-class dataset, our model is better than all other prototype-based models and comparable to black-box models. This confirms the robustness of our model observed in binary classification and also shows that our model produces accuracy comparable or better than black-box model even when the task is more challenging. Comparing our model against other prototype-based architectures, the performance is better in all the experiments done so far.

The consistent results across five datasets support our hypothesis in the beginning that GAProtoNet can perform on par with black-box and outperform prototype baselines. The performance improvement demonstrated by GAProtoNet indicates graph attention for prototypical networks can achieve better overall performance, thus validating our contribution.

\section{Interpretability Analysis} \label{4}
In this section, we analyze our model's interpretability from two perspectives. First, we use case studies to demonstrate how prototypes and their surrounding edges to the input node, constructed with graph attention, are used to explain the model's decision-making process. Second, we evaluate the quality of our prototypes. We demonstrate their representativeness through visualization, showing that they are distributed dispersedly within the training data space. Additionally, we assess the distinctiveness of the prototypes by examining how it varies with the number of prototypes and analyzing its impact on the model's prediction performance.
\begin{figure}[t]
\centering
  \includegraphics[width=0.9\columnwidth]{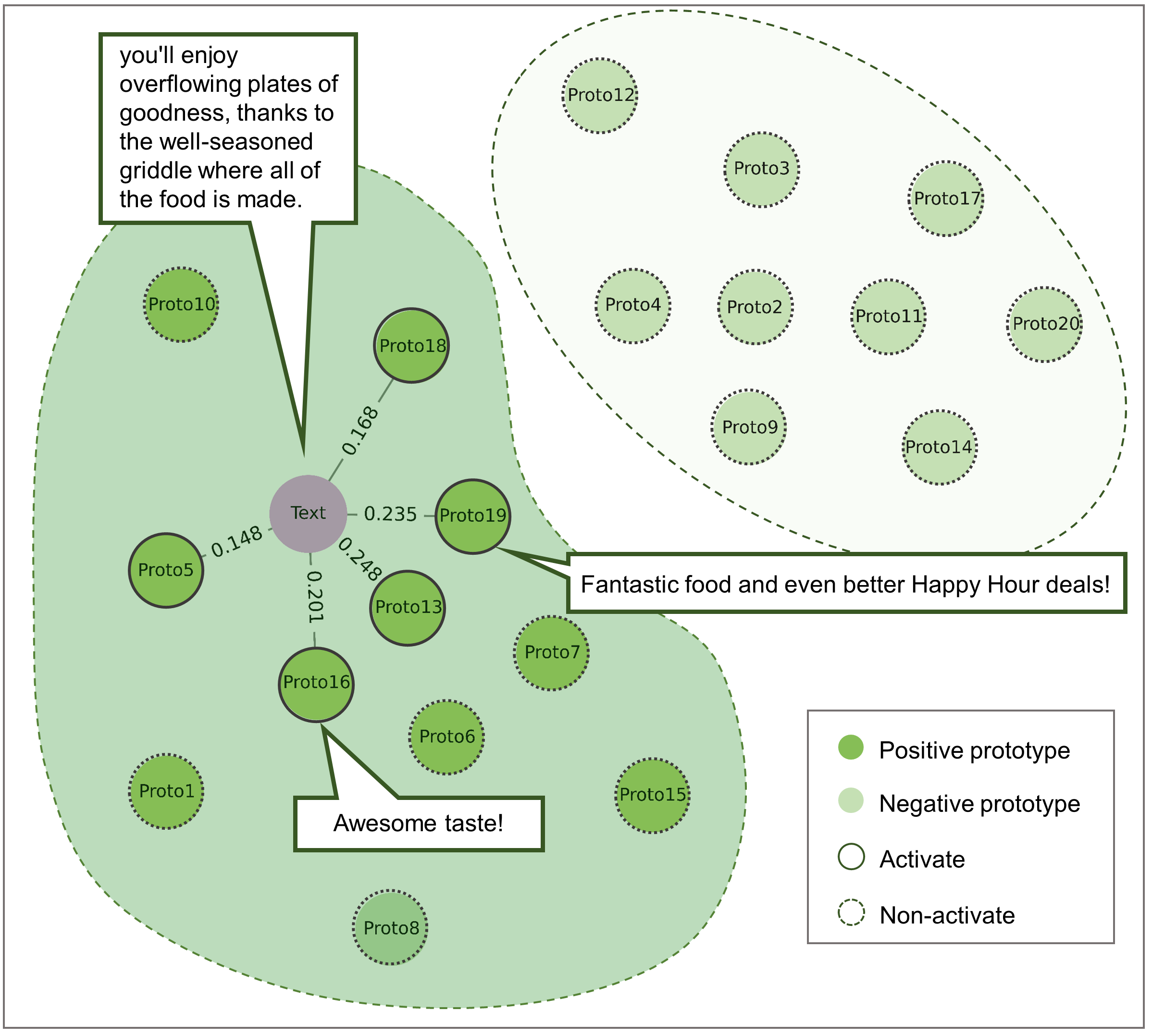}
  \caption{ Prototype activation under Attention Head 3. 
  % Dark green indicates positive prototypes while light green indicates negative prototypes. A solid outline around a circle denotes a prototype is activated.
  }
  \label{fig:graph}
\end{figure}

\begin{figure*}[t]
  \includegraphics[width=\textwidth]{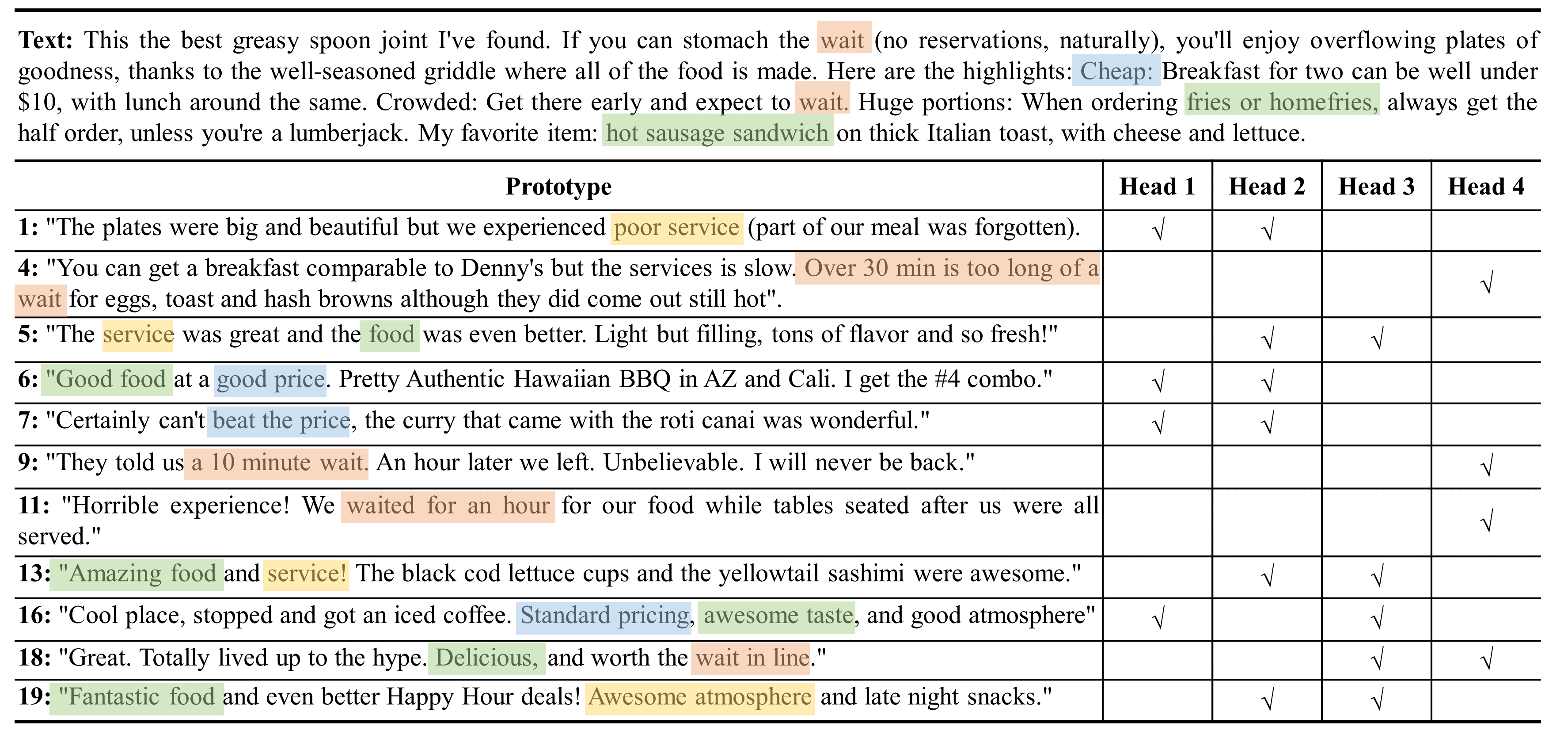} 
  \caption {Using a Yelp review as input, the prototypes activated by various graph attention heads are analyzed. The review's label and prediction are both positive. For each prototype, an attention head is checked if it constructs an edge between the prototype and the input. Different aspects are highlighted in distinct colors: blue for price, yellow for service, red for wait time, and green for food quality.}
    \label{case}
\end{figure*}

\begin{figure*}[h!]
    \centering
    \begin{minipage}[b]{0.3\textwidth}
        \centering
        \includegraphics[width=\textwidth]{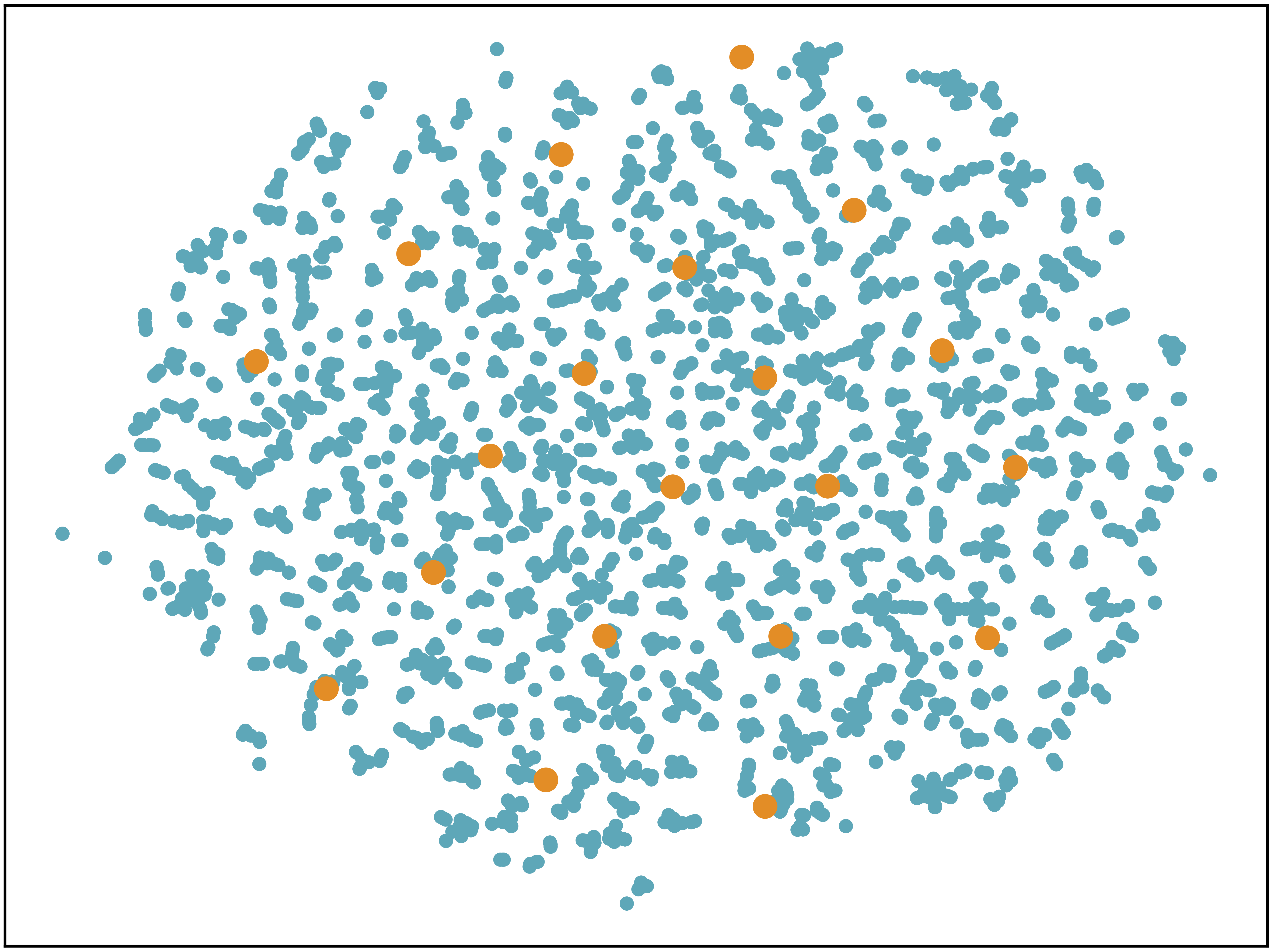}
        \subcaption{Hotel}

    \end{minipage}
    \hspace{0.02\textwidth}
    \begin{minipage}[b]{0.3\textwidth}
        \centering
        \includegraphics[width=\textwidth]{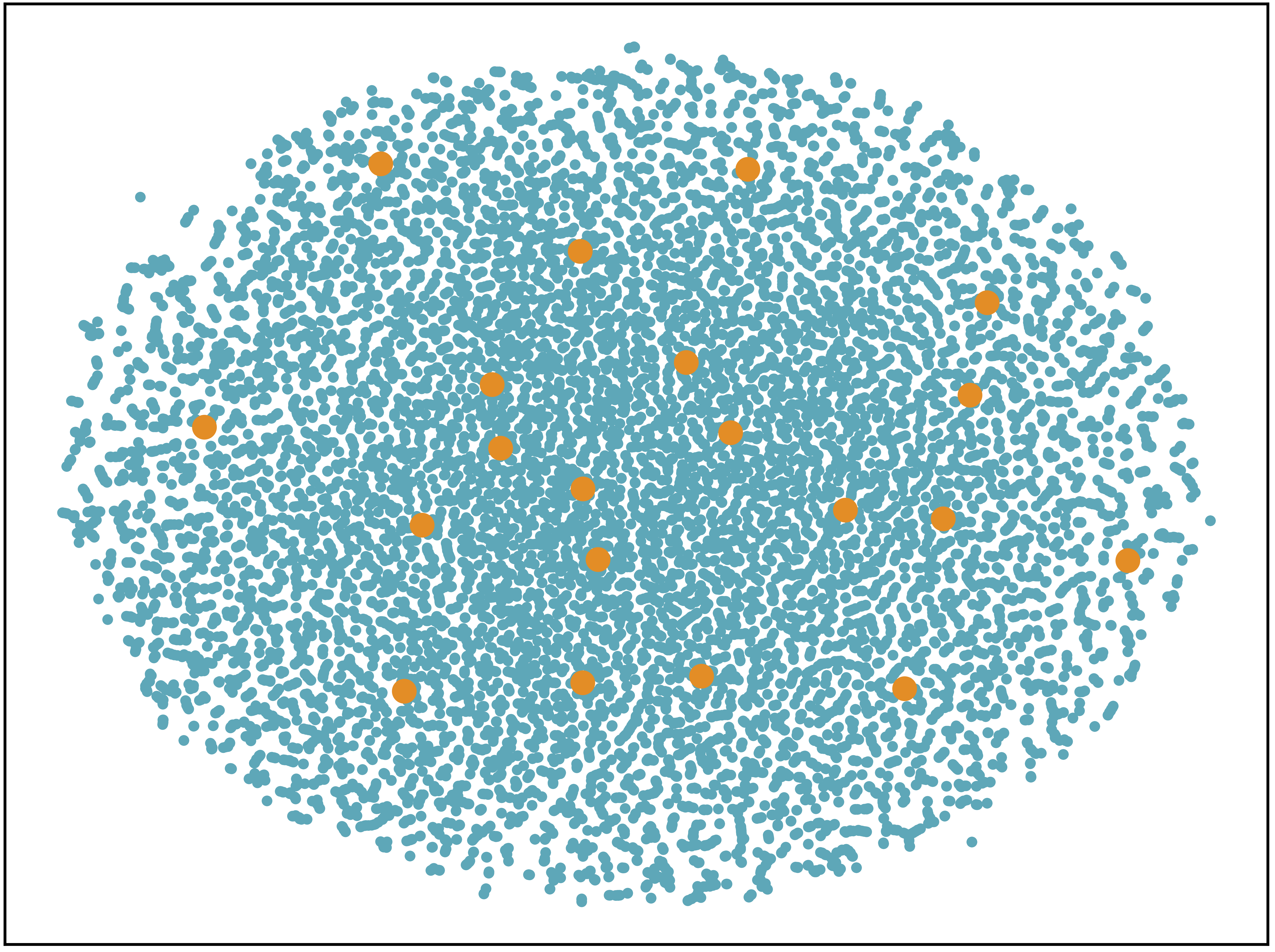}
        \subcaption{IMDb}
    \end{minipage}
    \hspace{0.02\textwidth}
    \begin{minipage}[b]{0.3\textwidth}
        \centering
        \includegraphics[width=\textwidth]{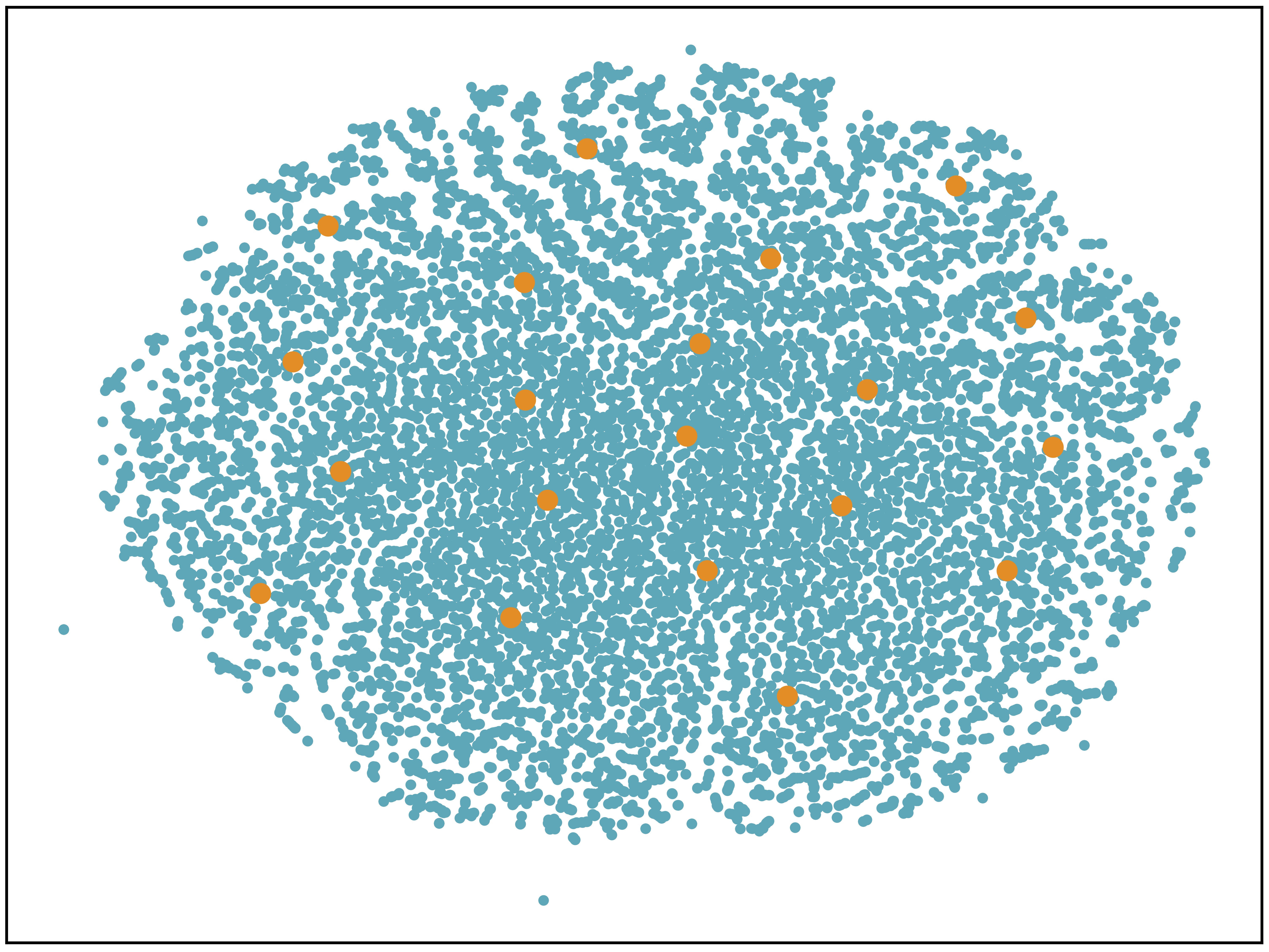}
        \subcaption{Yelp}
    \end{minipage}
    \caption{Prototype distribution within training data space for three datasets.}
    \label{fig:visualization}
\end{figure*}
\subsection{Case study}

Figure \ref{case} illustrates an example of using multi-graph attention heads and prototypes to explain a classification result. The input text is a Yelp review about a restaurant. In this instance, eleven out of twenty prototypes are activated by four attention heads, indicating edges constructed between them and the input text. Figure \ref{fig:graph} illustrates the activation of prototypes using attention Head 3 as an example. Specifically, five positive prototypes related to food quality are activated, and edges are constructed between the input text and the activated prototypes, with the attention scores representing the edge weights. This activation reflects the positive sentiment towards food quality expressed in the original text. 

We observe that different prototypes, representing various aspects of the restaurant, are activated by different attention heads. Head 1 activates prototypes related to service and price, Head 2 to service, Head 3 to food, and Head 4 to waiting time. In this text input, all different aspects of the text are captured by each attention head and then construct the edge between the input and the prototypes. Interestingly, prototypes related to service are activated by Head 2, even though the input text does not explicitly mention service. This suggests that the model might infer a positive attitude towards service based on implicit cues in the text. It also indicates that the aspects activated by different attention heads can overlap rather than being fully distinct.

\subsection{Prototype quality and performance}
\paragraph{Prototype Distribution Visualization} Since prototypes are used to interpret the model, they should capture as many representative features from the data as possible, indicating an even distribution within the training data space. A straightforward way to verify this is through visualization. Since our prototype vectors and training data are high-dimension vectors, we first use t-SNE (t-distributed Stochastic Neighbor Embedding) \cite{van2008visualizing} to do dimension deduction before the visualization. 
% It works by converting similarities between data points to joint probabilities and minimizing the Kullback-Leibler divergence \cite{kullback1951information} between the joint probabilities of the low-dimensional embedding and the high-dimensional data.
This method is effective at preserving the local structure of the data so the distribution pattern won't change when projected from higher dimension space to lower dimension space. The visualization of prototype vectors for a multi-head GAProtoNet with the prototype number of  $k=20$ is shown in Figure \ref{fig:visualization}. The orange dots represent prototypes while the blue dots represent training data. We can see that the prototype vectors are evenly distributed within the training data space and test results show that those prototype vectors are of high orthogonality. This indicates that the space formed by these vectors can cover most of the data points so we can use limited prototypes to represent any point in the training data.

\begin{figure}[t]
  \includegraphics[width=\columnwidth]{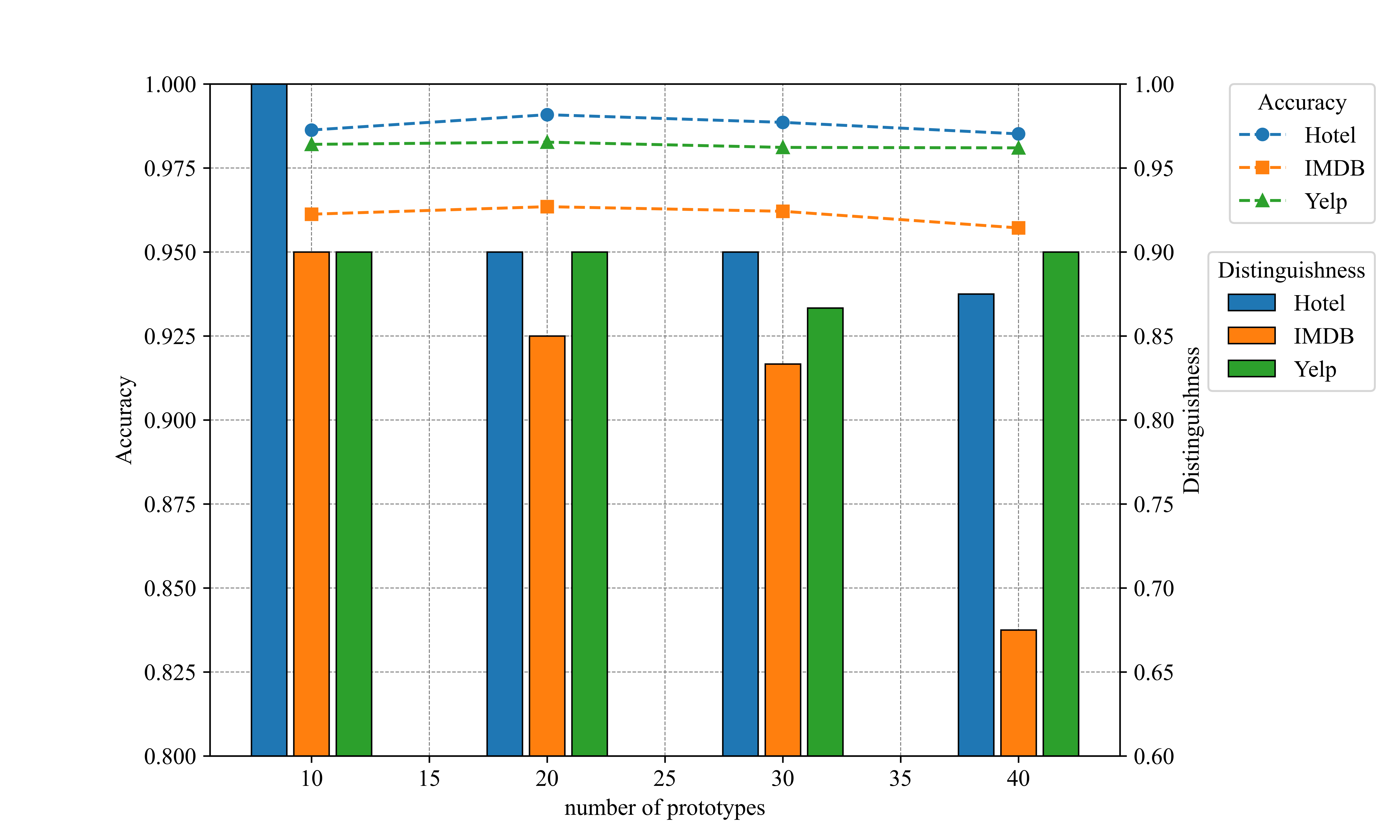}
  \caption{Prediction accuracy and the percentage of unique prototypes when trained with different numbers of prototypes }
  \label{fig:prototype}
\end{figure}

\paragraph{Prototype Distinguishness and Model Performance}
Since the prototypes are used both for interpreting results and making decisions, we are interested in how the hyperparameter $k$ of their number affects the model's interpretability and classification performance. For interpretability, we measure the number of distinguished prototypes. \textbf{\textit{Distinguishness}} indicates that if more than one prototype vector projects onto the same sample data point, it is counted only once. For classification performance, we measure the accuracy.  We conduct experiments varying the number of prototypes $k$ from $10$ to $40$ on a multi-head GAProtoNet with RoBERTa-large as the encoder across the three datasets. The percentage of distinguished prototypes and classification 
 accuracy is shown in Figure \ref{fig:prototype}. 
 % Overall, the percentage of distinguished prototypes are above $95\%$ and the accuracy is above $90\%$ except for the result on IMDb when $k=40$. 
 We observe the accuracy achieves the highest when $k=20$ across three datasets then drops with $k$ increasing. The corresponding trend for the percentage of distinguished vectors is also dropping from  $k=20$ to  $k=40$. We reason that increasing $k$ can only improve the model's expressiveness until a certain point, in our case $k=20$. After the point, prototypes will lose distinguishness and therefore hurts model's classification performance. Instead of observing a trade-off between interpretability and accuracy, our experiments show that they are positively related with each other.

\section{Related Work}\label{5}

  \citet{li2018deep} and \citet{chen2019looks} introduce prototype-based architecture into interpretable image classification by proposing a prototype layer, where prototypes are randomly initialized and are made meaningful through training. \citet{ming2019interpretable} brought this approach into NLP domain by adding a sequence encoder before the prototype layer. Concept representation learning is also applied in enhancing the interpretability of the structure \cite{jin-etal-2024-impact, jin2024exploring, jin2024prollm} while others improve interpretability by extracting minimal sentiment triplets \cite{sun-etal-2024-minicongts, sun2024enhancingnursingelderlycare}. However, despite this intrinsically interpretable model demonstrating compelling results, there are still some performance gaps compared to the original black-box model.

 In the NLP domain, researchers attempt to minimize the performance gap by trying different structure variations. \citet{hong2020protorynet} used Sentence Universal Encoder \cite{cer2018universal} and added an LSTM layer \cite{hochreiter1997long} between the prototype layer and the output layer to better capture the patterns in the trajectory of prototypes. \citet{plucinski2021prototypical} propose a new structure that operates on the prototypes in the form of phrases. \citet{das2022prototex} applied a new distance measurement metric and further minimized the accuracy gap but they only compared with the black-box models and lack of comparison with other prototype-based variations. 

All the above approaches utilize variations of heuristic distance metric (e.g. cosine similarity or Euclidean distance) to calculate the distance vector or matrix between the input vectors and prototypes, which serves as the sole input to the output layer. We hypothesize that only passing a heuristic distance metric may omit essential upstream information, thereby impairing performance. This motivates us to improve the current structure by calculating relatedness with the attention mechanism, which could potentially preserve more information \cite{vaswani2017attention}. 

% We are inspired by the work from \citet{arik2020protoattend}, in which they utilized \textit{relational attention} to determine the prototype weights by relating the input and the candidate samples via alignment of their keys and queries. However, since their overall approach is the traditional prototypical network stated in the beginning \cite{snell2017prototypical}, during each training step all candidate samples in the training dataset need to be fed into the encoder and do backpropagation, which leads to a significantly longer training time. Moreover, the size of the candidate database is constrained by the processor's memory capacity.

Considering all the limitations of the current approaches, there is a need for a new prototypical network that can further minimize the performance gap compared with the original black-box model and at the same time prototypes can be effectively trained through active learning.

% This line of research can be categorized into two broad directions. One popular group of approaches is to use surrogate models and post-hoc explainability techniques, which are separate from the original decision-making process. For example, Local Interpretable Model-agnostic Explanations (LIME) \cite{ribeiro2016should} and Shapley Additive Explanations (SHAP) \cite{lundberg2017unified} are used to generate explanations for text classification model by assigning weight scores to each input word representation\cite{bedi2021multi}. However, recent research  (\citealp{rudin2019stop}, \citealp{alvarez2018robustness}) has pointed out that these methods face fundamental limitations as post-hoc explanation techniques: they suffer from unfaithfulness and inconsistency in the explanations since they only approximate the decision-making process rather than being the actual decision-makers. Additionally, various studies have shown that these models are sensitive to small perturbations in the target model’s inputs or parameters (\citealp{ivankay2022fooling}, \citealp{mardaoui2021analysis}).

\section{Conclusion} \label{6}
We contribute an interpretable prototypical deep learning architecture that is advanced with graph attention on text classification. Our experiments show our approach outperforms both other interpretable and black-box architectures on benchmark datasets. We also conduct a comprehensive analysis to show good interpretability of our approach.

\section{Limitation} \label{7}
To demonstrate these prototypes can support user explainability in addition to making the architecture interpretable, we will include user study in our future work.

% Bibliography entries for the entire Anthology, followed by custom entries
%\bibliography{anthology,custom}
% Custom bibliography entries only
\bibliography{acl_latex}

\begin{thebibliography}{27}
\providecommand{\natexlab}[1]{#1}

\bibitem[{Bhatti et~al.(2023)Bhatti, Huang, Neira-Molina, Marjan, Baryalai, Tang, Wu, and Bazai}]{bhatti2023mffcg}
Uzair~Aslam Bhatti, Mengxing Huang, Harold Neira-Molina, Shah Marjan, Mehmood Baryalai, Hao Tang, Guilu Wu, and Sibghat~Ullah Bazai. 2023.
\newblock Mffcg--multi feature fusion for hyperspectral image classification using graph attention network.
\newblock \emph{Expert Systems with Applications}, 229:120496.

\bibitem[{Cer et~al.(2018)Cer, Yang, Kong, Hua, Limtiaco, John, Constant, Guajardo-Cespedes, Yuan, Tar et~al.}]{cer2018universal}
Daniel Cer, Yinfei Yang, Sheng-yi Kong, Nan Hua, Nicole Limtiaco, Rhomni~St John, Noah Constant, Mario Guajardo-Cespedes, Steve Yuan, Chris Tar, et~al. 2018.
\newblock Universal sentence encoder.
\newblock \emph{arXiv preprint arXiv:1803.11175}.

\bibitem[{Chen et~al.(2019)Chen, Li, Tao, Barnett, Rudin, and Su}]{chen2019looks}
Chaofan Chen, Oscar Li, Daniel Tao, Alina Barnett, Cynthia Rudin, and Jonathan~K Su. 2019.
\newblock This looks like that: deep learning for interpretable image recognition.
\newblock \emph{Advances in neural information processing systems}, 32.

\bibitem[{Das et~al.(2022)Das, Gupta, Kovatchev, Lease, and Li}]{das2022prototex}
Anubrata Das, Chitrank Gupta, Venelin Kovatchev, Matthew Lease, and Junyi~Jessy Li. 2022.
\newblock Prototex: Explaining model decisions with prototype tensors.
\newblock \emph{arXiv preprint arXiv:2204.05426}.

\bibitem[{Datta and Kibler(1995)}]{datta1995learning}
Piew Datta and Dennis Kibler. 1995.
\newblock Learning prototypical concept descriptions.
\newblock In \emph{Machine Learning Proceedings 1995}, pages 158--166. Elsevier.

\bibitem[{Devlin et~al.(2018)Devlin, Chang, Lee, and Toutanova}]{devlin2018bert}
Jacob Devlin, Ming-Wei Chang, Kenton Lee, and Kristina Toutanova. 2018.
\newblock Bert: Pre-training of deep bidirectional transformers for language understanding.
\newblock \emph{arXiv preprint arXiv:1810.04805}.

\bibitem[{Hochreiter and Schmidhuber(1997)}]{hochreiter1997long}
Sepp Hochreiter and J{\"u}rgen Schmidhuber. 1997.
\newblock Long short-term memory.
\newblock \emph{Neural computation}, 9(8):1735--1780.

\bibitem[{Hong et~al.(2020)Hong, Wang, and Baek}]{hong2020protorynet}
Dat Hong, Tong Wang, and Stephen~S Baek. 2020.
\newblock Protorynet-interpretable text classification via prototype trajectories.
\newblock \emph{arXiv preprint arXiv:2007.01777}.

\bibitem[{Jin et~al.(2024{\natexlab{a}})Jin, Xue, Wang, Kang, Ye, Zhou, Du, and Zhang}]{jin2024prollm}
Mingyu Jin, Haochen Xue, Zhenting Wang, Boming Kang, Ruosong Ye, Kaixiong Zhou, Mengnan Du, and Yongfeng Zhang. 2024{\natexlab{a}}.
\newblock Prollm: Protein chain-of-thoughts enhanced llm for protein-protein interaction prediction.
\newblock \emph{arXiv e-prints}, pages arXiv--2405.

\bibitem[{Jin et~al.(2024{\natexlab{b}})Jin, Yu, Huang, Zeng, Wang, Hua, Zhao, Mei, Meng, Ding et~al.}]{jin2024exploring}
Mingyu Jin, Qinkai Yu, Jingyuan Huang, Qingcheng Zeng, Zhenting Wang, Wenyue Hua, Haiyan Zhao, Kai Mei, Yanda Meng, Kaize Ding, et~al. 2024{\natexlab{b}}.
\newblock Exploring concept depth: How large language models acquire knowledge at different layers?
\newblock \emph{arXiv preprint arXiv:2404.07066}.

\bibitem[{Jin et~al.(2024{\natexlab{c}})Jin, Yu, Shu, Zhao, Hua, Meng, Zhang, and Du}]{jin-etal-2024-impact}
Mingyu Jin, Qinkai Yu, Dong Shu, Haiyan Zhao, Wenyue Hua, Yanda Meng, Yongfeng Zhang, and Mengnan Du. 2024{\natexlab{c}}.
\newblock \href {https://aclanthology.org/2024.findings-acl.108} {The impact of reasoning step length on large language models}.
\newblock In \emph{Findings of the Association for Computational Linguistics ACL 2024}, pages 1830--1842, Bangkok, Thailand and virtual meeting. Association for Computational Linguistics.

\bibitem[{Li et~al.(2018)Li, Liu, Chen, and Rudin}]{li2018deep}
Oscar Li, Hao Liu, Chaofan Chen, and Cynthia Rudin. 2018.
\newblock Deep learning for case-based reasoning through prototypes: A neural network that explains its predictions.
\newblock In \emph{Proceedings of the AAAI Conference on Artificial Intelligence}, volume~32.

\bibitem[{Liu et~al.(2019)Liu, Ott, Goyal, Du, Joshi, Chen, Levy, Lewis, Zettlemoyer, and Stoyanov}]{liu2019roberta}
Yinhan Liu, Myle Ott, Naman Goyal, Jingfei Du, Mandar Joshi, Danqi Chen, Omer Levy, Mike Lewis, Luke Zettlemoyer, and Veselin Stoyanov. 2019.
\newblock \href {https://arxiv.org/abs/1907.11692} {Roberta: A robustly optimized bert pretraining approach}.
\newblock \emph{Preprint}, arXiv:1907.11692.

\bibitem[{Maas et~al.(2011)Maas, Daly, Pham, Huang, Ng, and Potts}]{maas-EtAl:2011:ACL-HLT2011}
Andrew~L. Maas, Raymond~E. Daly, Peter~T. Pham, Dan Huang, Andrew~Y. Ng, and Christopher Potts. 2011.
\newblock \href {http://www.aclweb.org/anthology/P11-1015} {Learning word vectors for sentiment analysis}.
\newblock In \emph{Proceedings of the 49th Annual Meeting of the Association for Computational Linguistics: Human Language Technologies}, pages 142--150, Portland, Oregon, USA. Association for Computational Linguistics.

\bibitem[{Ming et~al.(2019)Ming, Xu, Qu, and Ren}]{ming2019interpretable}
Yao Ming, Panpan Xu, Huamin Qu, and Liu Ren. 2019.
\newblock Interpretable and steerable sequence learning via prototypes.
\newblock In \emph{Proceedings of the 25th ACM SIGKDD International Conference on Knowledge Discovery \& Data Mining}, pages 903--913.

\bibitem[{Mohammad et~al.(2018)Mohammad, Bravo-Marquez, Salameh, and Kiritchenko}]{mohammad-etal-2018-semeval}
Saif Mohammad, Felipe Bravo-Marquez, Mohammad Salameh, and Svetlana Kiritchenko. 2018.
\newblock \href {https://doi.org/10.18653/v1/S18-1001} {{S}em{E}val-2018 task 1: Affect in tweets}.
\newblock In \emph{Proceedings of the 12th International Workshop on Semantic Evaluation}, pages 1--17, New Orleans, Louisiana. Association for Computational Linguistics.

\bibitem[{Pluci{\'n}ski et~al.(2021)Pluci{\'n}ski, Lango, and Stefanowski}]{plucinski2021prototypical}
Kamil Pluci{\'n}ski, Mateusz Lango, and Jerzy Stefanowski. 2021.
\newblock Prototypical convolutional neural network for a phrase-based explanation of sentiment classification.
\newblock In \emph{Joint European Conference on Machine Learning and Knowledge Discovery in Databases}, pages 457--472. Springer.

\bibitem[{Sanh et~al.(2019)Sanh, Debut, Chaumond, and Wolf}]{sanh2019distilbert}
Victor Sanh, Lysandre Debut, Julien Chaumond, and Thomas Wolf. 2019.
\newblock Distilbert, a distilled version of bert: smaller, faster, cheaper and lighter.
\newblock \emph{arXiv preprint arXiv:1910.01108}.

\bibitem[{Sun et~al.(2024{\natexlab{a}})Sun, Xie, Ye, Gu, and Guo}]{sun2024enhancingnursingelderlycare}
Qiao Sun, Jiexin Xie, Nanyang Ye, Qinying Gu, and Shijie Guo. 2024{\natexlab{a}}.
\newblock \href {https://arxiv.org/abs/2412.09946} {Enhancing nursing and elderly care with large language models: An ai-driven framework}.
\newblock \emph{Preprint}, arXiv:2412.09946.

\bibitem[{Sun et~al.(2024{\natexlab{b}})Sun, Yang, Ma, Ye, and Gu}]{sun-etal-2024-minicongts}
Qiao Sun, Liujia Yang, Minghao Ma, Nanyang Ye, and Qinying Gu. 2024{\natexlab{b}}.
\newblock \href {https://doi.org/10.18653/v1/2024.emnlp-main.165} {{M}ini{C}on{GTS}: A near ultimate minimalist contrastive grid tagging scheme for aspect sentiment triplet extraction}.
\newblock In \emph{Proceedings of the 2024 Conference on Empirical Methods in Natural Language Processing}, pages 2817--2834, Miami, Florida, USA. Association for Computational Linguistics.

\bibitem[{Van~der Maaten and Hinton(2008)}]{van2008visualizing}
Laurens Van~der Maaten and Geoffrey Hinton. 2008.
\newblock Visualizing data using t-sne.
\newblock \emph{Journal of machine learning research}, 9(11).

\bibitem[{Vaswani et~al.(2017)Vaswani, Shazeer, Parmar, Uszkoreit, Jones, Gomez, Kaiser, and Polosukhin}]{vaswani2017attention}
Ashish Vaswani, Noam Shazeer, Niki Parmar, Jakob Uszkoreit, Llion Jones, Aidan~N Gomez, {\L}ukasz Kaiser, and Illia Polosukhin. 2017.
\newblock Attention is all you need.
\newblock \emph{Advances in neural information processing systems}, 30.

\bibitem[{Velickovic et~al.(2017)Velickovic, Cucurull, Casanova, Romero, Lio, Bengio et~al.}]{velickovic2017graph}
Petar Velickovic, Guillem Cucurull, Arantxa Casanova, Adriana Romero, Pietro Lio, Yoshua Bengio, et~al. 2017.
\newblock Graph attention networks.
\newblock \emph{stat}, 1050(20):10--48550.

\bibitem[{Wang et~al.(2019)Wang, He, Cao, Liu, and Chua}]{wang2019kgat}
Xiang Wang, Xiangnan He, Yixin Cao, Meng Liu, and Tat-Seng Chua. 2019.
\newblock Kgat: Knowledge graph attention network for recommendation.
\newblock In \emph{Proceedings of the 25th ACM SIGKDD international conference on knowledge discovery \& data mining}, pages 950--958.

\bibitem[{Xie et~al.(2020)Xie, Zhang, Gong, Tang, and Han}]{xie2020mgat}
Yu~Xie, Yuanqiao Zhang, Maoguo Gong, Zedong Tang, and Chao Han. 2020.
\newblock Mgat: Multi-view graph attention networks.
\newblock \emph{Neural Networks}, 132:180--189.

\bibitem[{Yang et~al.(2019)Yang, Dai, Yang, Carbonell, Salakhutdinov, and Le}]{yang2019xlnet}
Zhilin Yang, Zihang Dai, Yiming Yang, Jaime Carbonell, Russ~R Salakhutdinov, and Quoc~V Le. 2019.
\newblock Xlnet: Generalized autoregressive pretraining for language understanding.
\newblock \emph{Advances in neural information processing systems}, 32.

\bibitem[{Zhang et~al.(2015)Zhang, Zhao, and LeCun}]{zhang2015character}
Xiang Zhang, Junbo Zhao, and Yann LeCun. 2015.
\newblock Character-level convolutional networks for text classification.
\newblock \emph{Advances in neural information processing systems}, 28.

\end{thebibliography}
\begin{figure*}[t]
  \includegraphics[width=\textwidth]{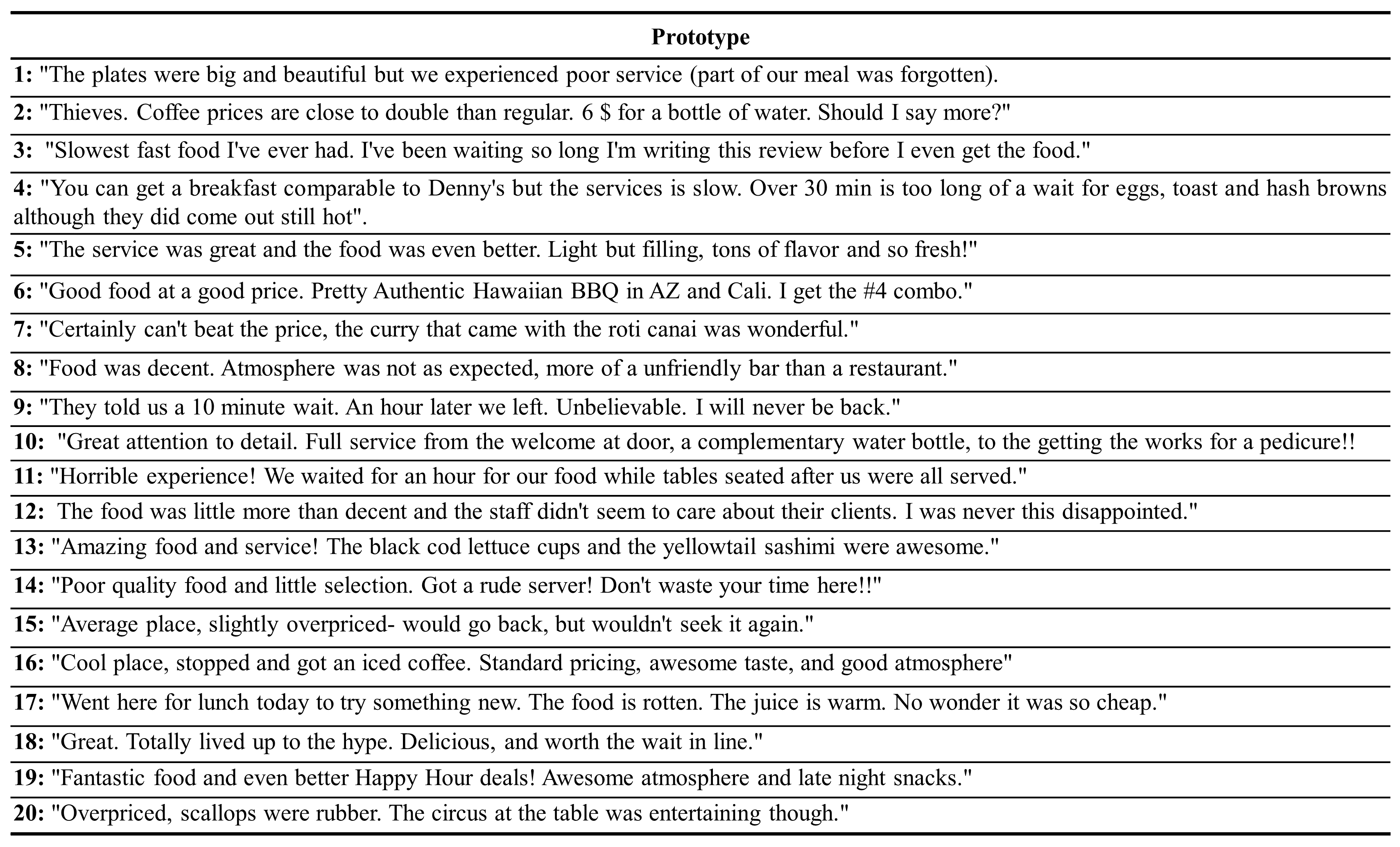} 
  \caption {This is the 20 prototypes after projecting them with the nearest sample.}
    \label{prototypes} 
\end{figure*}
\appendix

\section{Projected Prototypes}
\label{sec:appendix}

\end{document}